\documentclass[10pt,twocolumn,letterpaper]{article}

\usepackage{cvpr}              
\usepackage{array}

\usepackage{siunitx}
\definecolor{cvprblue}{rgb}{0.21,0.49,0.74}
\usepackage[pagebackref,breaklinks,colorlinks,allcolors=cvprblue]{hyperref}

\newcommand{\submissionnumber}{4}

\def\paperID{\submissionnumber} 
\def\confName{CVPR}
\def\confYear{2026}

\def\methodname{MakeupMirror\xspace}
\def\datasetname{MakeupSelfies\xspace}

\newcommand{\fig}[1]{Fig.~\ref{fig:#1}}
\newcommand{\sect}[1]{Sect.~\ref{sec:#1}}

\title{\methodname: Improving Facial Attribute Preservation \\ in Diffusion Models for Makeup Transfer}

\author{
Nefeli Andreou,
Angel Martínez-González,
Sabine Sternig,
Matthieu Guillaumin, \\
Epameinondas Antonakos and
Michael Opitz \\
Amazon \\
{\tt\small \{nandreou, gonamang, ssternig, matthieg, antonak, micopitz\}@amazon.de }
}

\begin{document}
\maketitle

\begin{abstract}
Makeup transfer models enable fun augmented reality~(AR) experiences as well as virtual try-on~(VTO) for online makeup shopping. While recent state-of-the-art diffusion-based solutions such as Stable-Makeup~\cite{stable_makeup:2025} dramatically improve the accuracy and realism of makeup transfer, they still face limitations in identity and skin color preservation, making production-level VTO for makeup shopping unrealistic. In this work, we propose \methodname, a diffusion-based approach to makeup transfer that makes significant progress towards preserving facial features and skin tone. We introduce several technical innovations over Stable-Makeup:
\begin{enumerate*}[(\arabic*)]

\item integration of facial geometry conditioning with ControlNets to maintain facial fidelity;

\item region-specific makeup transfer control to enable precise makeup application
across facial regions such as skin, eyes and lips;

\item skin tone-based makeup transfer modulation that prevent skin tone
alteration in cross-subject transfer scenarios; and 
\item integration of a Levenberg-Marquardt Langevin sampler to speed up inference while maintaining generation quality.
\end{enumerate*}
Our experiments on CPM-Real, Makeup Wild, and (herein newly collected, more diverse) \datasetname datasets show that \methodname{} improves relative facial recognition similarity by $+60\%$, reduces relative skin tone difference by $-50\%$ over Stable-Makeup, with a latency of 0.7s, while achieving expert acceptance rate of 94\% across core facial identity preservation criteria.

\end{abstract}

\begin{figure}
    \centering
    \includegraphics[width=1\linewidth]{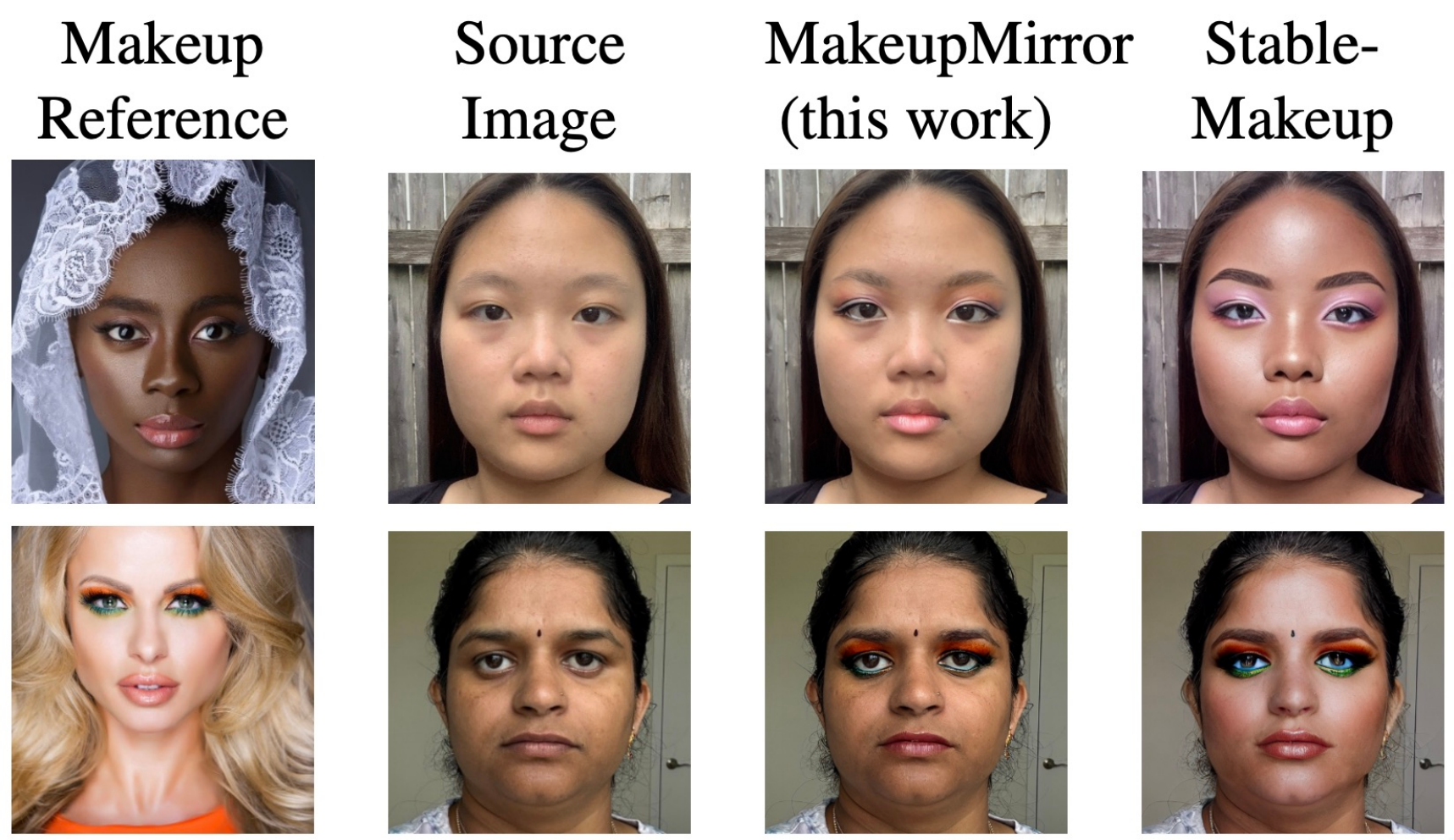}
    \caption{\textbf{Makeup transfer, qualitative comparison.}
Makeup transfer is the task of applying a reference makeup (left image) to a source input image (second column).
While both methods achieve photorealism and faithful makeup transfer, \methodname{} (third column) better preserves facial attributes and skin tone of the source image compared to Stable-Makeup~\cite{stable_makeup:2025} (right image).}
\label{fig:intro_fig}
\end{figure}

\section{Introduction}
\label{sec:intro}
\emph{Makeup transfer} is the task of virtually applying the visible set of makeup products of a reference face image to a source face, while preserving its identity, see Fig.~\ref{fig:intro_fig}.
Recent progress in the field has enabled several real-world applications, such as entertaining augmented reality (AR) functions, and is narrowing the gap to hyper-realistic virtual try-on (VTO) systems for beauty e-commerce, with which customers could faithfully visualize how products will render on their own unique features before they buy products.
To fully close that gap, makeup transfer models still face  challenges in appearance understanding and generation:
makeup properties and application styles must be effectively disentangled from facial identity, transferred across subjects with different lighting, pose, expression, skin tones and facial geometries, and applied with region specific control while maintaining photorealistic quality.

Traditional physically-based rendering approaches to makeup transfer~\cite{Tong2007ExampleBasedCT,Li:2019, Liu:2024, Lin:2013, Guo2009DigitalFM} are fast and offer precise control but struggle to achieve photorealistic results, particularly for complex makeup looks involving multiple products of various textures.
Early works in the generative AI space explored GAN-based approaches~\cite{jiang2020psgan,yang2022elegant} which offer improved quality but face inherent instability during training and lack full customization and controllability in makeup application.
Following the image generation and editing domains, recent advances in makeup transfer leverage diffusion models~\cite{stable_makeup:2025,park2025dreammakeup,sun2024shmt}, given their better controllability and improved visual fidelity.
However, existing diffusion-based methods still face important limitations: unintended modifications to facial features and skin tones, as well as exaggerated and unfaithful transfer, see~\fig{intro_fig}.
These technical shortcomings -- in particular unintentional skin tone, eye or nose shape modifications -- render current approaches unsuitable for real-world e-commerce applications where fidelity, photorealism, speed, and controllability are all essential.

In this work, we introduce \methodname{}, a diffusion-based model making significant progress in preserving facial attributes of the source image.
Our improvements enable faithful makeup transfer, allowing users to apply looks with photorealistic quality, yet also speed, controllability, as well as convincing facial feature and skin tone preservation.
Our approach builds upon Stable-Makeup~\cite{stable_makeup:2025}, the state-of-the-art diffusion-based architecture for makeup transfer and makes the following contributions:
\begin{enumerate}
    \item \textbf{Enhanced Facial Fidelity through Geometric Conditioning:} We integrate ControlNets~\cite{zhang2023adding} for Depth-Anything~\cite{yang2024depth} estimation and Canny edge~\cite{cannyedge} detection maps to maintain facial structure, low-level details and overall identity during makeup transfer;
    \item \textbf{Region-Specific Makeup Strength Control:} We implement adaptive control mechanisms which reduce classifier-free guidance and diffusion steps for skin regions while maintaining larger transfer for lips and eyes.
    This enables precise makeup application in facial regions where it is prominent while preventing facial feature transfer in other regions;
    \item \textbf{Adaptive Skin Tone Preservation:} We incorporate skin tone difference detection between the reference and source images so as to automatically modulate makeup transfer intensity.
    When significant variations exist, this modulation prevents unwanted skin tone modifications while preserving makeup accuracy.
    \item \textbf{Inference Acceleration:} We integrate a Levenberg–Marquardt Langevin sampler to accelerate inference, achieving a $2.8\times$ speed-up while maintaining makeup transfer quality, leading to a latency of 0.7s.
\end{enumerate}

In the remainder of the paper, we first discuss related work (\sect{related_work}) then present our contributions in detail (\sect{method}).
In our experiments (\sect{experiments}) on CPM-Real~\cite{m_Nguyen-etal-CVPR21}, Makeup Wild datasets~\cite{jiang2020psgan}, as well as on a newly collected one displaying larger diversity, which we denote as \datasetname, we show that \methodname{} improves relative facial recognition similarity by $+60\%$ and reduces relative skin tone difference by $-50\%$ compared to Stable-Makeup and leads to a 94\% pass-rate in an audit conducted by beauty experts.
We also conduct an ablation study and show how recent advances in diffusion sampling~\cite{wang2025unleashing} speed-up \methodname{} by $2.8\times$ without significant impact on quality.

\section{Related Work}
\label{sec:related_work}

\subsection{Diffusion Models}
Diffusion models are a powerful family of generative models that synthesize high-quality images through an iterative denoising process.
Following DDPM~\cite{ho2020denoising} which demonstrated the feasibility of recovering structured data from Gaussian noise, the extension to conditional
diffusion enabled significant 
progress in numerous generation tasks including
text-to-image~\cite{podellsdxl,ramesh2022hierarchical,rombach2022high,saharia2022photorealistic},
text-to-video~\cite{Blattmann2023StableVD, kling2025klingomni, wan2024videogen},
inpainting~\cite{lugmayr2022repaint}, and
image editing~\cite{Epstein:2023,Li:2023,zhang2023sine,xie2023edit,tsaban2023ledits}.
In this setting, the model predicts noise conditioned on an external signal, typically injected 
via cross-attention, introducing semantic control over the generative process while preserving the 
core denoising formulation.

Early models using conditional diffusion struggled with computational efficiency and accurate
adherence to the controlling signals.
Latent Diffusion Models (LDM)~\cite{rombach2022high} address the high computational cost of 
pixel-space diffusion by performing the denoising process in a compressed latent representation
of the image computed with image encoders.
Classifier guidance~\cite{dhariwal2021diffusion} first introduced gradient-based 
conditioning using classifiers predictions from noisy images.
Classifier-Free Guidance~\cite{ho2021classifier} later introduced a sampling-time technique that strengthens prompt 
adherence by linearly combining conditional and unconditional noise predictions without an external 
classifier.
In particular, Stable-Diffusion~\cite{rombach2022high} builds on the LDM design at scale integrating 
latent diffusion, cross-attention conditioning, and classifier-free guidance.
Several other architectural extensions such as ControlNets~\cite{zhang2023adding}, 
T2I-Adapters~\cite{mou2023t2i}, and LoRA-based conditining~\cite{yu2023lora}
further introduced fine-grained structured spatial control, using signals such as pose, depth, edges, and layout, while preserving pre-trained priors.
More recently, preference-based finetuning~\cite{wallace2024diffusiondpo} and 
reinforcement learning~\cite{wu2022unified} 
approaches further improved preference alignment beyond prompt application.

\begin{figure*}[t]
    \centering
    \includegraphics[width=0.9\linewidth]{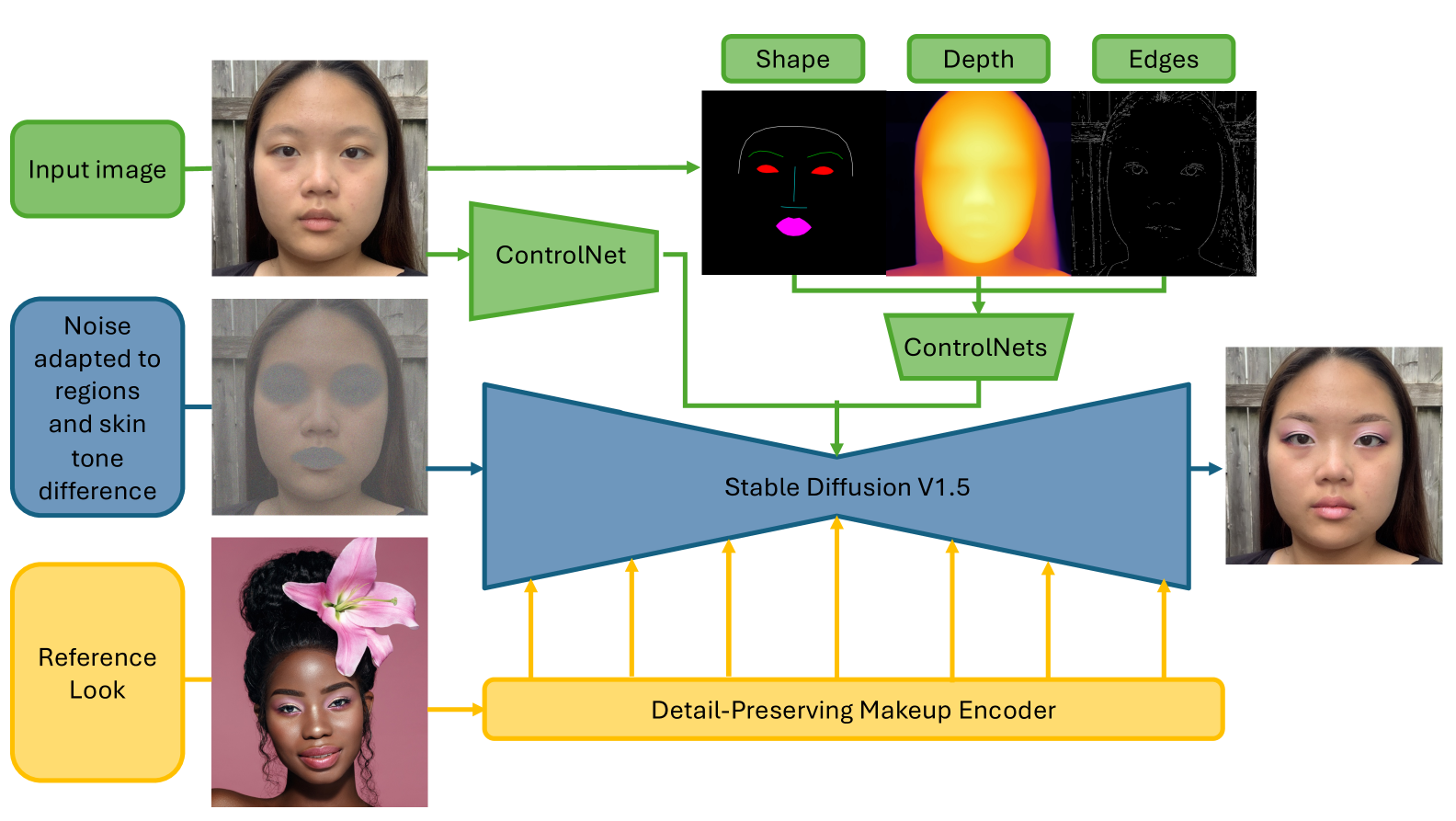}
    \caption{\textbf{Illustration of the \methodname{} architecture.} Building upon Stable-Makeup, our approach enhances facial feature preservation by adding controller networks for Depth-Anything~\cite{yang2024depth} and Canny edge~\cite{cannyedge} maps. We also adapt noise to facial segmentation regions and modulate the transfer strength based on the estimated skin tone difference between the source image and the reference look.}
    \label{fig:method}
\end{figure*}

\subsection{Makeup Transfer}
Traditional makeup transfer relied on image processing and graphics techniques, including intrinsic decomposition and physically-based rendering~\cite{Tong2007ExampleBasedCT,Li:2019, Liu:2024, Lin:2013, Guo2009DigitalFM}.
Later, works in the space of generative AI removed explicit 3D world modeling and
instead used a Generative Adversarial Networks (GAN) 
trained on large datasets of (\textit{makeup}, \textit{non-makeup}) image 
pairs to learn to transfer makeup~\cite{yang2022elegant,Xiang:2022,jiang2020psgan,gu2019ladn,Li:2018}.
Despite their large adoption, GAN-based methods lack diversity representation and often need facial
region alignment, preventing their practical application in real-world scenarios and complex 
makeup compositions.

Building on recent image generation improvements, most works now leverage diffusion models for the makeup transfer task~\cite{lu2024makeupdiffuse, pan2026supervised, zhu2025flux, park2025dreammakeup, stable_makeup:2025}.
A core challenge addressed in these methods is the disentanglement of person identity and
global makeup representation.
The approach presented in~\cite{lu2024makeupdiffuse} integrates conditional diffusion with an adversarial 
discriminator that regularizes the denoising process, improving realism while preserving identity 
through spatially aligned makeup conditioning. In~\cite{sun2024shmt} the authors leverage 3D facial information to guide the diffusion process and improve
facial geometric consistency while learning makeup representations in a self-supervised manner, addressing
the lack of paired image data for training. Zhang \etal~\cite{stable_makeup:2025} introduced Stable-Makeup, a framework that performs makeup transfer by
conditioning on facial pose to ensure preservation of facial identity while keeping makeup accuracy.
In their method, makeup and pose representations are combined within the U-Net architecture using spatial-aware 
cross attention, allowing makeup features to be applied consistently across corresponding facial 
regions, e.g. eyes, lips, etc.
More recently, FLUX-Makeup~\cite{zhu2025flux} alleviates the need for face-control modules by injecting region-aware makeup style information via LoRA modules trained on before-after makeup image pairs.

These methods exhibit undesirable failures in real-world cross-subject scenarios. In our experiments we observed that they often unintentionally alter facial features and skin tones. We argue that these limitations are likely inherited from their reliance on same-subject paired training data. 
Our work addresses these fundamental limitations by introducing a novel approach that preserves subject identity while achieving accurate makeup transfer. 
We  incorporate facial geometric features and skin tone information into the diffusion process to explicitly address issues arising in cross-subject scenarios.
Furthermore, we introduce region-specific strength control for precise makeup application, which not only enhances transfer accuracy but also enables efficient processing by reducing the diffusion steps.

\section{Method}
\label{sec:method}
In this section, we present our enhanced makeup transfer approach, as illustrated in~\fig{method}.
We first describe the foundational architecture that builds upon Stable-Makeup~\cite{stable_makeup:2025} as the basis of our method (Sec.~\ref{sec:background}), then detail our technical innovations:
geometric conditioning with ControlNets~\cite{zhang2023adding} (Sec.~\ref{sec:conditioning}), region-specific makeup strength control (Sec.~\ref{sec:skin_control}) and adaptive skin tone preservation (Sec.~\ref{sec:skin_detection}).

\subsection{Background}
\label{sec:background}
Stable-Makeup~\cite{stable_makeup:2025} is the de facto state-of-the-art model for makeup transfer, itself building upon Stable Diffusion V1-5, which employs a LDM framework with a U-Net denoising architecture. Starting from noise $\mathbf{z_T} \sim \mathcal{N}(\mathbf{0}, \mathbf{I})$ sampled from a standard Gaussian distribution, the model performs a series of $T$ iterative denoising steps to recover a clean latent $\mathbf{z_0}$. At each timestep $t \in \{T, T-1, \ldots, 1\}$, the denoising U-Net $\epsilon_\theta$ predicts the noise component $\epsilon_\theta(\mathbf{z}_t, t, \mathbf{c})$, where $\mathbf{c}$ encodes the conditioning signals (input image and reference makeup). The sequential denoising process is guided via classifier-free guidance~\cite{ho2021classifier}, which interpolates between conditional and unconditional noise predictions to control the fidelity–diversity tradeoff. 

The Detail-Preserving (D-P) makeup encoder in Stable-Makeup extracts multi-scale makeup representations from reference images to capture intricate details such as eyelashes, eyebrows and fine lines.
Formally, the encoder processes a reference makeup image $I_m$ through a pretrained CLIP~\cite{Radford2021LearningTV} visual backbone to produce multi-scale detailed makeup embeddings $E_m = \mathrm{concat}_{k=0}^K(E_k, \mathrm{dim}\!=\!1)$, where $E_k$ represents image embeddings at layer $k$ in the CLIP visual backbone.
The makeup embeddings are fed through a self-attention layer, which better preserves the multi-layer features compared to a linear layer, and incorporated into the U-Net via cross-attention layers.
To maintain consistency with the source image, Stable-Makeup incorporates two adapted ControlNet~\cite{zhang2023adding} encoders: the content encoder processes the source image $I_s$ to preserve pixel-level content and non-facial regions while the structural encoder uses dense colored lines based on facial keypoints to preserve facial structure.

\subsection{Enhanced Geometric Conditioning}
\label{sec:conditioning}
While the baseline Stable-Makeup architecture demonstrates effective makeup transfer capabilities, we observe undesired behaviors including unintentional skin tone modifications and facial feature alterations during the diffusion process (see~\fig{intro_fig}).
To address these limitations and maintain facial fidelity throughout makeup application, we employ two additional pretrained ControlNets that provide explicit and complementary geometric constraints to the diffusion U-Net.
The depth conditioning utilizes the efficient Depth-Anything model~\cite{yang2024depth} to predict depth maps $F_D$ from the source image $I_s$, thus encoding three-dimensional facial structure information that preserves geometric relationships during the generation process at low computational cost.
In parallel, a Canny edge~\cite{cannyedge} map $F_E$ captures fine-grained contours including facial feature edges, jawlines, and significant anatomical landmarks.
Both feature maps are injected into the corresponding layers of the main U-Net decoder through additive connections (\cf upper part of~\fig{method}).

\subsection{Region-Specific Strength Control}
\label{sec:skin_control}
Existing makeup transfer methods using generative AI apply uniform transfer intensity across all facial regions, failing to disentangle makeup products from underlying facial characteristics, which results in over-application and unnatural skin tone modifications especially when the input and reference makeup image have significantly different skin tones. We implement region-specific makeup strength control that modulates transfer intensity based on facial segmentation, enabling differential treatment of skin regions versus feature regions (lips and eyes) within a single forward pass of the diffusion model. For segmentation, we handcraft a template mask and warp it on the input image.

Our approach operates by dynamically adjusting two key diffusion parameters: the classifier-free guidance scale $w_\text{c}$ and the number of diffusion time-steps $T_\text{c}$.
We define a template makeup transfer mask $M$ which delineates facial regions where makeup should be applied, enabling spatial modulation of the diffusion process.
The template mask is warped into the source image using a piecewise affine transformation computed from facial landmarks~\cite{mediapipe}.
During the denoising process, the classifier-free guidance scale, $w_\text{c}$, controls the strength of makeup conditioning for skin regions $\tilde{\epsilon}_\theta(\textbf{z}_t,t,\textbf{c}) = (1+w_\text{c}) \epsilon_\theta(\textbf{z}_t,t, \textbf{c}) - w_c\epsilon_\theta(\textbf{z}_t,t,\emptyset)$ by interpolating between conditional and unconditional predictions~\cite{ho2021classifier}, while the clamping time-steps parameter $T_c$ constrains when makeup conditioning is applied. As seen in Fig.~\ref{fig:method} the mask allows for higher noise to by applied on the lips and eyes, and introduces lower noise to the skin area.
For time-steps $t > T_\text{c}$, the diffusion process operates without makeup conditioning on skin regions, effectively reducing transfer intensity while maintaining full conditioning for feature regions throughout all time-steps.

We define three discrete transfer strength levels $s \in \{\text{low}, \text{medium}, \text{high}\}$ with parameters $(w_c(s), T_c(s))$, where $w_c$ denotes the classifier-free guidance scale for skin regions and $T_c$ denotes the clamping time-steps:
$$
(w_c(s), T_c(s)) =
\begin{cases}
(1.05, \lfloor 0.93 \times T \rfloor) & \text{if } s = \text{low} \\
(1.1, \lfloor 0.87 \times T \rfloor) & \text{if } s = \text{medium} \\
(1.6, \lfloor 0.73 \times T \rfloor) & \text{if } s = \text{high}
\end{cases}
$$
where $T$ is the total number of inference steps.

\subsection{Adaptive Skin Tone Preservation}
\label{sec:skin_detection}
Makeup transfer between individuals with significantly different skin tones presents a fundamental challenge: aggressive transfer can alter the recipient's natural skin tone, while conservative transfer may fail to capture the intended makeup aesthetic.
We introduce adaptive skin tone difference detection that automatically modulates transfer intensity based on the perceptual skin tone difference between source and reference images.

Our system computes skin tone differences using the Monk scale~\cite{monk_2023}, a perceptually-uniform 10-point skin tone classification system. 
The Monk scale values are extracted through a linear classifier trained on CLIP embeddings. 
Given source image $I_s$ and reference makeup image $I_m$, we extract their respective Monk scale classifications $\mu_s$ and $\mu_m$ in the range $[1, \ldots, 10]$.
The skin tone difference is computed as the absolute difference: $\Delta{\mu} = |\mu_s - \mu_m|$.
When significant skin tone variations are detected ($\Delta{\mu} \ge 4$), the system reduces the transfer strength by one level.

\begin{figure*}[htbp]
\centering
\begin{subfigure}{0.49\linewidth}
\centering
\includegraphics[width=0.96\linewidth]{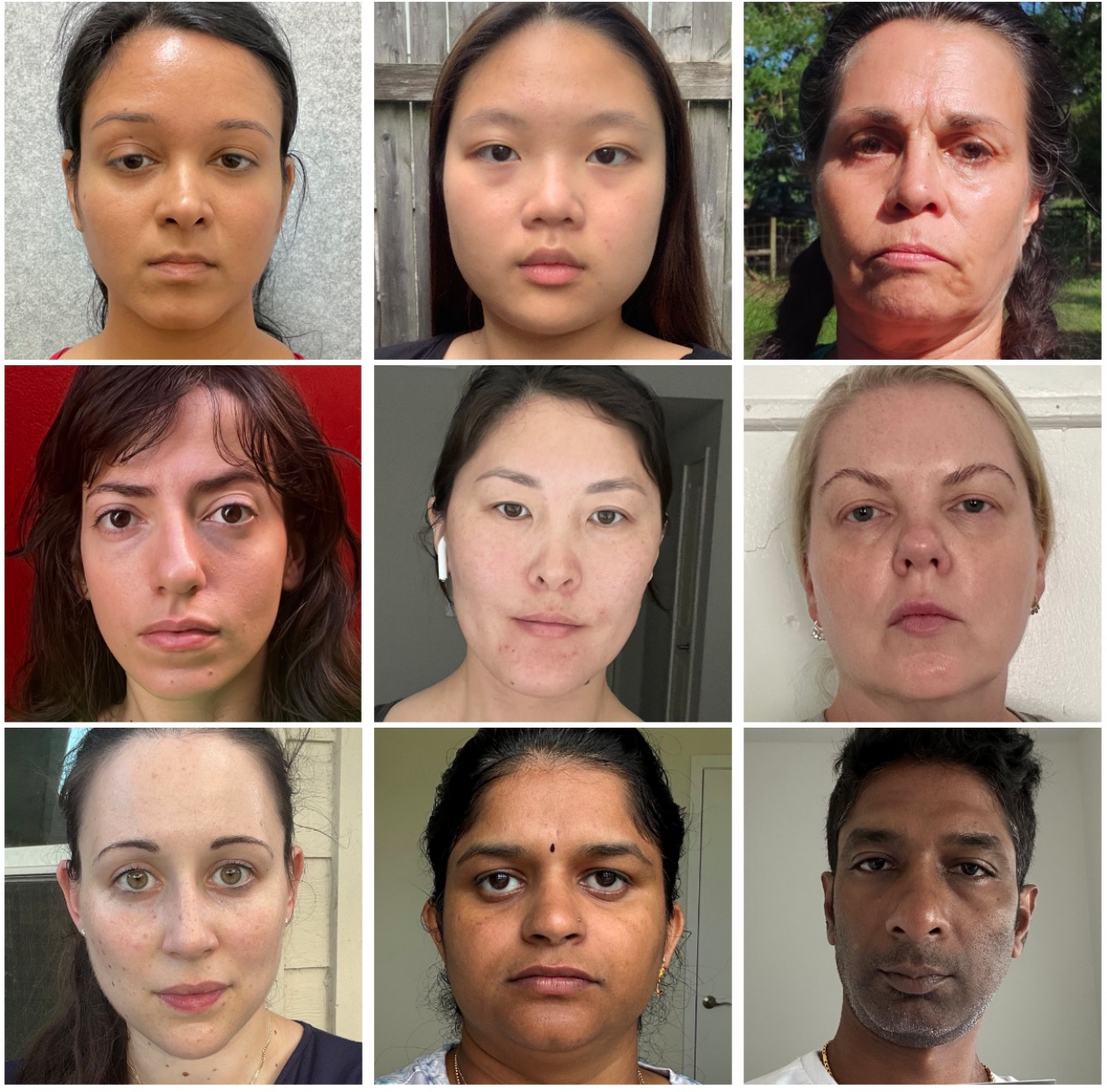}
\caption{Sample of selfies without makeup in \datasetname.}
\label{fig:selfies-qualitative}
\end{subfigure}
\begin{subfigure}{0.49\linewidth}
\includegraphics[width=0.96\linewidth]{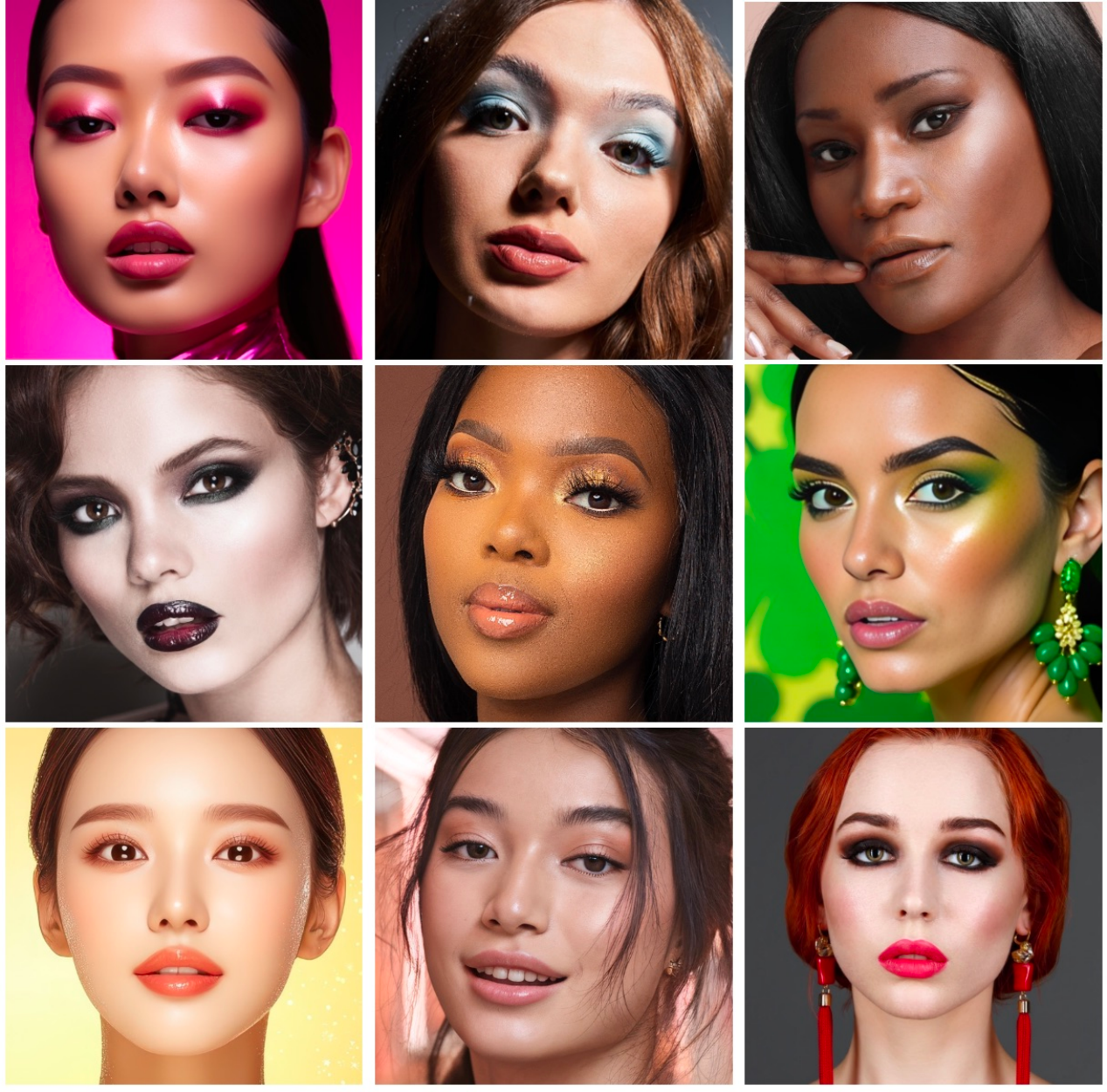}
\caption{Sample of stock images used as reference looks in \datasetname.}
\label{fig:stock-images}
\end{subfigure}
\caption{Sample images (source images and reference makeup image) from the herein collected \datasetname dataset.}
\end{figure*}

\section{Experiments}
\label{sec:experiments}
In this section, we provide a thorough evaluation of our method.
Public datasets, \eg~\cite{jiang2020psgan,Li:2018,m_Nguyen-etal-CVPR21}, tend to lack diversity in skin tone, raising bias and social fairness concerns.
Consequently, we collect a new dataset with a wider variety of skin tones consisting of selfie images, serving as non-makeup source images, as well as curated stock images, serving as reference makeup images (Sec.~\ref{sec:dataset-collection}).
We show that our method achieves significantly better identity preservation compared to the state-of-the-art on public and our private benchmarks, while preserving the makeup on makeup-specific regions (Sec.~\ref{sec:sota-comparison}).
Next, we show a detailed ablation study of our design choices (Sec.~\ref{sec:ablation-study}).
Finally, we show how we optimize the latency of our method for real-world deployment (Sec.~\ref{sec:latency}).

\subsection{\datasetname Dataset Collection}
\label{sec:dataset-collection}

To enable a fair and bias-free evaluation of our method which mimics an online shopping use-case, we curate two sets of imagery, which we denote as \datasetname.

First, we recorded a set of images of faces without makeup, serving as input images.
The goal is to recreate the in-store try-on experience, enabling customers to virtually try on products from the comfort of their own home. To this end, we collected about $1k$ frontal-view selfies from smartphone cameras in indoor and outdoor environments from a diverse group of people, \cf samples in \fig{selfies-qualitative}.

Second, we collected a set of visually elevated images with makeup, serving as makeup reference images. To this end, we leveraged high-quality licensed stock images.
To collect these images, we used an LLM to generate $\approx\!\!100$ makeup related search queries.
Next, we crawled stock websites with these search queries to get about $50k$ initial thumbnail images.
However, these images can contain multiple humans, small faces, or non-frontal viewing faces.
Consequently, we used heuristics to identify frontal-view face close-up images and remove images with multiple faces.
Finally, as many stock images tend to show the same person, we used face-embeddings to de-duplicate identities, resulting in a shortlist of about $5.5k$ images.
To ensure diversity, we used a skin tone prediction model, to group these images into monk scale skin tone buckets.
Finally, human curators select a balanced set of 880 images from these skin tone buckets. Example images can be seen in Fig.~\ref{fig:stock-images}.

\begin{figure}[htbp]
\centering

\begin{subfigure}{\linewidth}
    \centering
    \includegraphics[width=\linewidth]{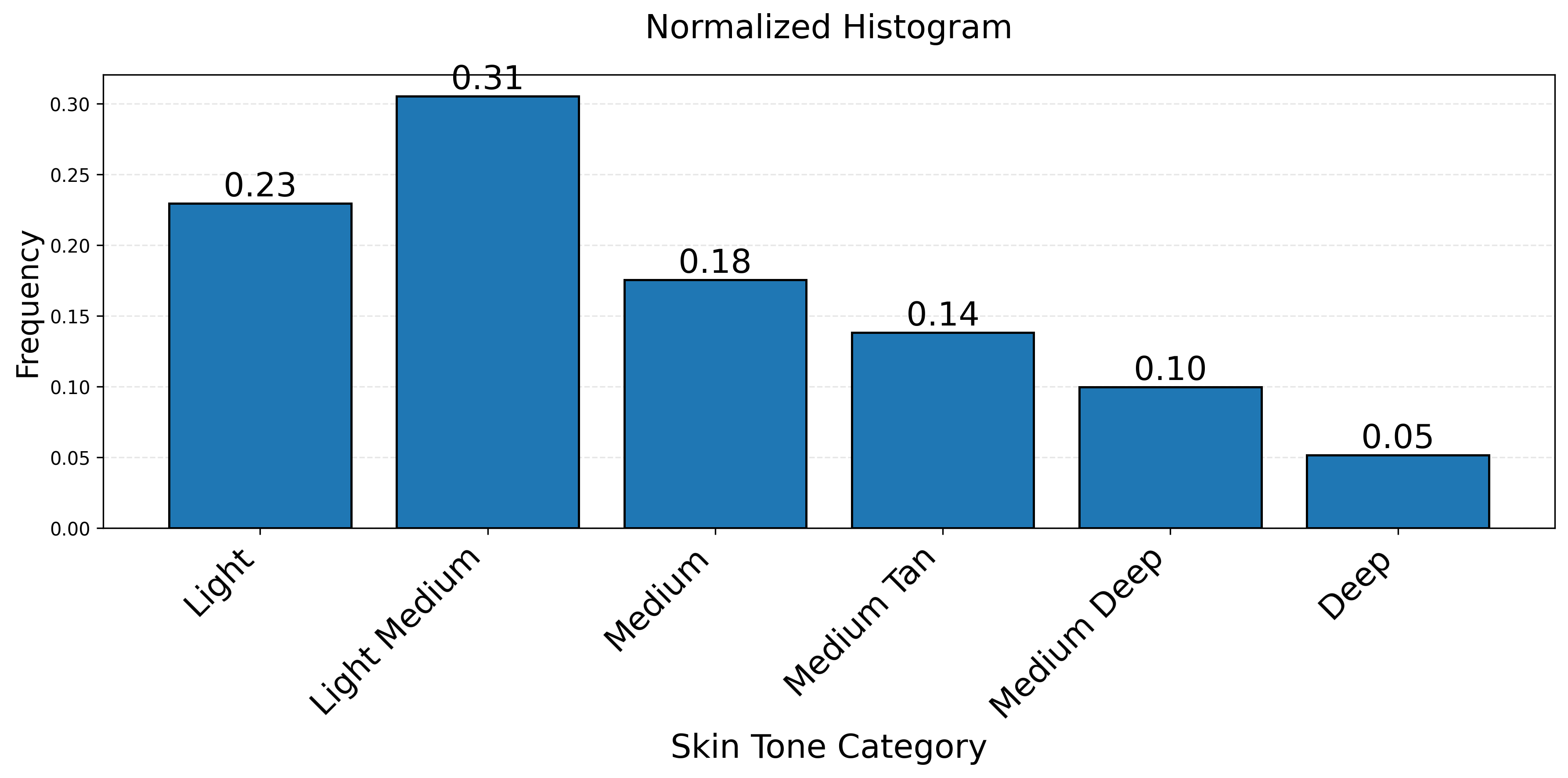}
    \caption{Skin tone distribution in our selfie image collection.}
    \label{fig:selfie-skin-distribution}
\end{subfigure}
\begin{subfigure}{\linewidth}
    \centering
    \includegraphics[width=\linewidth]{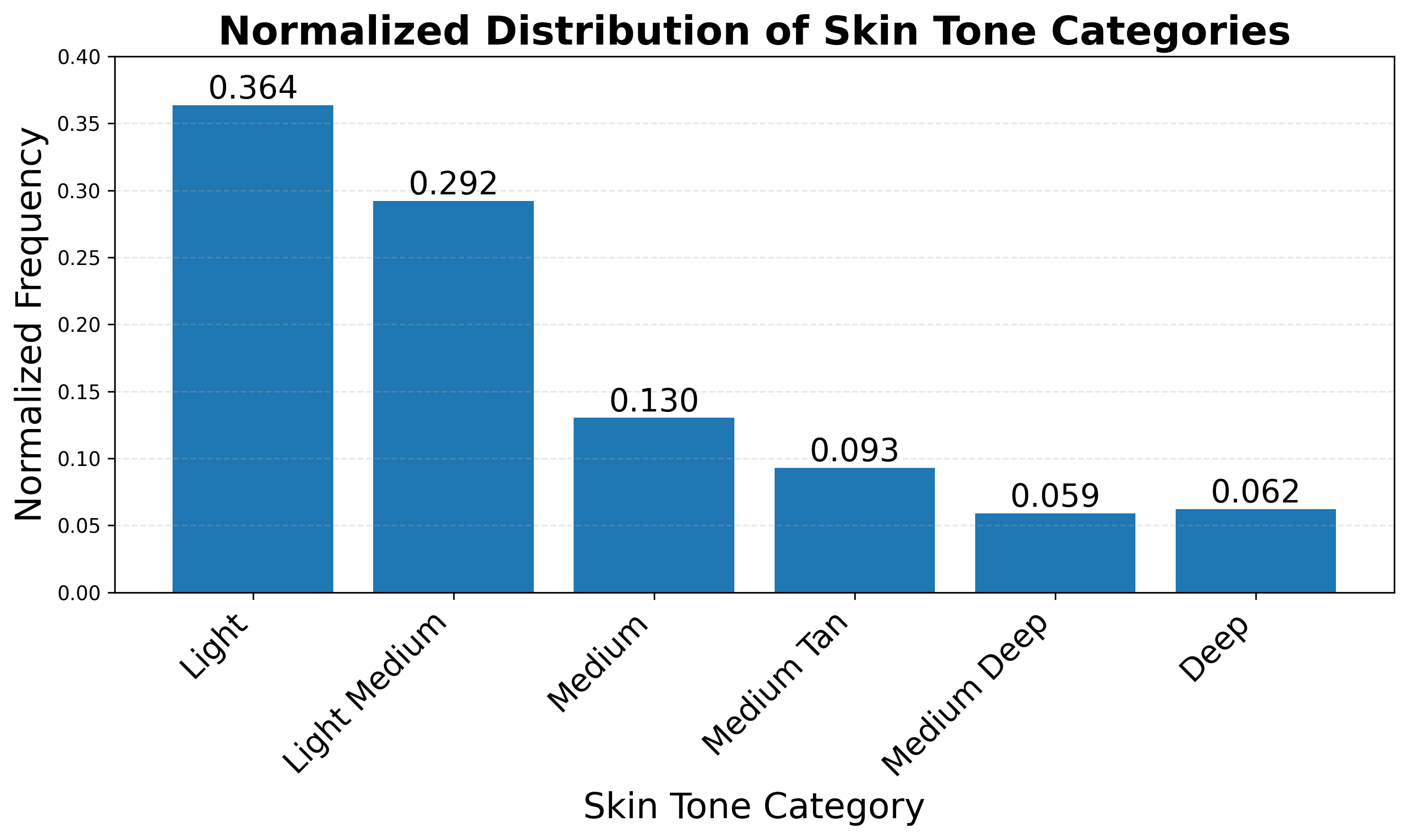}
    \caption{Skin tone distribution in our licensed stock image collection.}
    \label{fig:stock-skin-distribution}
\end{subfigure}
\begin{subfigure}{\linewidth}
    \centering
    \includegraphics[width=\linewidth]{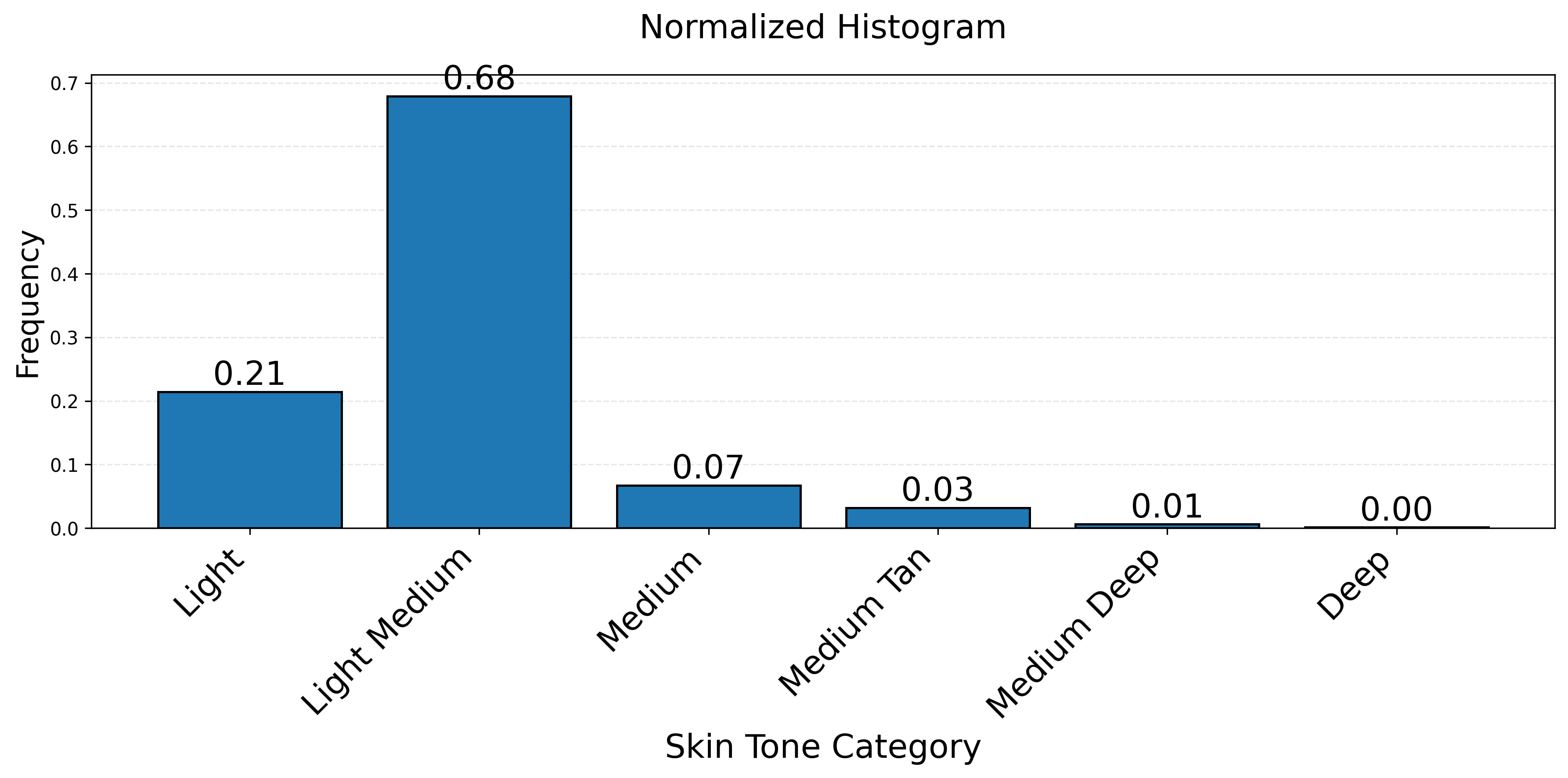}
    \caption{Skin tone distribution in public datasets~\cite{jiang2020psgan,Li:2018,m_Nguyen-etal-CVPR21}.}
    \label{fig:public-skin-distribution}
\end{subfigure}
\caption{Skin tone distributions of our datasets (top, middle) compared to existing public datasets (bottom). Our datasets have a more balanced skin tone distribution, enabling a fairer and bias-free evaluation.}
\end{figure}

To highlight the diversity of our dataset, we ran the same skin tone analysis on public datasets~\cite{jiang2020psgan,Li:2018,m_Nguyen-etal-CVPR21} and compare the skin tone distribution (\fig{public-skin-distribution}) with that of the selfies (\fig{selfie-skin-distribution}) and reference looks (\fig{stock-skin-distribution}).
As shown in \fig{public-skin-distribution}, these public datasets tend to predominantly display individuals in lighter skin buckets, making a bias-free evaluation virtually impossible, while \datasetname displays a more balanced distribution over skin tones in both subsets.

\subsection{Comparison to the State-of-the-Art}
\label{sec:sota-comparison}

In this section, we show that our method quantitatively and qualitatively outperforms the state-of-the-art StableMakeup~\cite{stable_makeup:2025} on two public datasets, \ie~CPM-Real~\cite{m_Nguyen-etal-CVPR21}, and Makeup Wild~\cite{jiang2020psgan}, as well as our \datasetname dataset.

To this end, we follow existing evaluation protocols~\cite{stable_makeup:2025}, and sample 2,000 makeup and non-makeup pairs for each of our datasets.
As CPM-Real does not have non-makeup images, we use the non-makeup images of the Makeup Transfer~\cite{Li:2018} dataset to construct these pairs. 
To evaluate on \datasetname, we follow a similar protocol, and sample non-makeup selfies with corresponding licensed stock makeup images for evaluation.

To measure identity preservation, we use a skin tone prediction model, as well as a face verification model~\cite{serengil2026boosted}.
More formally, given source image $I_s$, and transferred output image $I_o$, we extract their monk scale $\mu_s$ and $\mu_o$ and compute $\Delta \mu = |\mu_s - \mu_o|$.
The lower the monk scale difference, the more accurately the transfer method preserves the skin tone.
For face similarity, we extract FaceNet512 embeddings via DeepFace~\cite{serengil2026boosted}.
Similarly, the higher the similarity between the source image and transferred output image, the more we preserve the facial identity. 

To measure makeup fidelity, we leverage DINOv2~\cite{oquab2023dinov2} embeddings.
As there is an inherent trade-off in preserving facial identity and makeup fidelity, we compute these features on the full face (Dv2-F), as well as on makeup-specific regions (Dv2-M), \ie the eye region and mouth region.
More specifically, we measure similarity between makeup reference image $I_m$ and transferred makeup image $I_o$.
To extract makeup-specific regions, we warp our makeup transfer mask (Sec.~\ref{sec:skin_control}) on the makeup and output image, and compute DINOv2 features only on the non-masked regions (\ie eyes, lips).

We summarize our results in Table~\ref{tbl:evaluation}.
Compared to the baseline Stable-Makeup~\cite{stable_makeup:2025}, our method preserves facial identity significantly better across all datasets -- about $-50\%$ for $\Delta\!$~Monk, $+60\%$ for Face Sim, and flat for Dv2-M -- while preserving makeup fidelity in makeup-specific regions.
In regions such as cheeks, nose, forehead, \etc, our method compares unfavorably to our baseline in terms of DINOv2 similarity (Dv2-F).
We hypothesize that a high DINOv2 similarity in these regions typically also corresponds to saturated makeup and betrays unintentional skin tone change, and thus a loss in identity fidelity. Existing metrics conflate makeup coverage with accuracy, leaving precise, scalable makeup evaluation an open challenge.

Further, we illustrate qualitative results in \fig{qualitative-results}.
We see that our approach preserves the facial identity significantly better compared to Stable-Makeup.
To quantify any critical defects of our method, we also performed a manual human audit on about $2k$ samples pairs with 3 beauty expert annotators/sample.
Specifically, we focus on detecting (1) defects on the lip region, (2) defects in the eye region (\eg eye crease disappearing), (3) eyebrow quality (\eg duplicated eyebrows), and (4) change of skin tone.
Overall, we have found that our solution achieves a pass rate of 94\% across these defects.

\begin{table*}[t]
\setlength{\tabcolsep}{3pt}
\centering
\resizebox{\textwidth}{!}{
\begin{tabular}{l|cccc|cccc|cccc}
\hline
& \multicolumn{4}{|c|}{CPM-Real} & \multicolumn{4}{|c|}{Makeup Wild} & \multicolumn{4}{|c}{\datasetname} \\
\hline
Method & 
Dv2 (M) & Dv2 (F) & $\Delta$ Monk & Face Sim. & 
Dv2 (M) & Dv2 (F) & $\Delta$ Monk & Face Sim. & 
Dv2 (M) & Dv2 (F) & $\Delta$ Monk & Face Sim. 
\\
\hline
StableMakeup~\cite{stable_makeup:2025} & 
\textbf{0.418} & \textbf{0.614} & 0.701 & 0.523 &
0.249 & \textbf{0.631} & 0.741 & 0.590  &
\textbf{0.583} & \textbf{0.627} & 1.132 & 0.487
\\
\methodname{} & 
0.416 & 0.392 & \textbf{0.349} & \textbf{0.886} &
\textbf{0.251} & 0.503 & \textbf{0.320} & \textbf{0.893} &
0.5631 & 0.452 & \textbf{0.427} & \textbf{0.867} \\
\hline

\end{tabular}
}
\caption{Our method preserves facial identity significantly better compared to our baseline, as measured in monk scale difference ($\Delta$~Monk) and face similarity (Face Sim.). It preserves makeup on makeup-specific regions, as measured in DINOv2 similarity (Dv2~(M)). }
\label{tbl:evaluation}
\end{table*}

\subsection{Ablation Study}
\label{sec:ablation-study}

Here, we show the effectiveness of each of our contributions quantitatively and qualitatively on our \datasetname dataset.
In Table~\ref{tbl:ablation} we see that introducing additional depth and Canny ControlNets already improves facial identity preservation (+0.177) as well as monk scale difference (-0.277), while maintaining makeup fidelity in makeup-specific regions.
We achieve further improvements, when we add our novel skin tone preservation method, improving the face similarity (+0.203), and reducing the monk scale difference (-0.428).

\begin{figure*}[!htbp]
    \centering
    \includegraphics[width=1\linewidth]{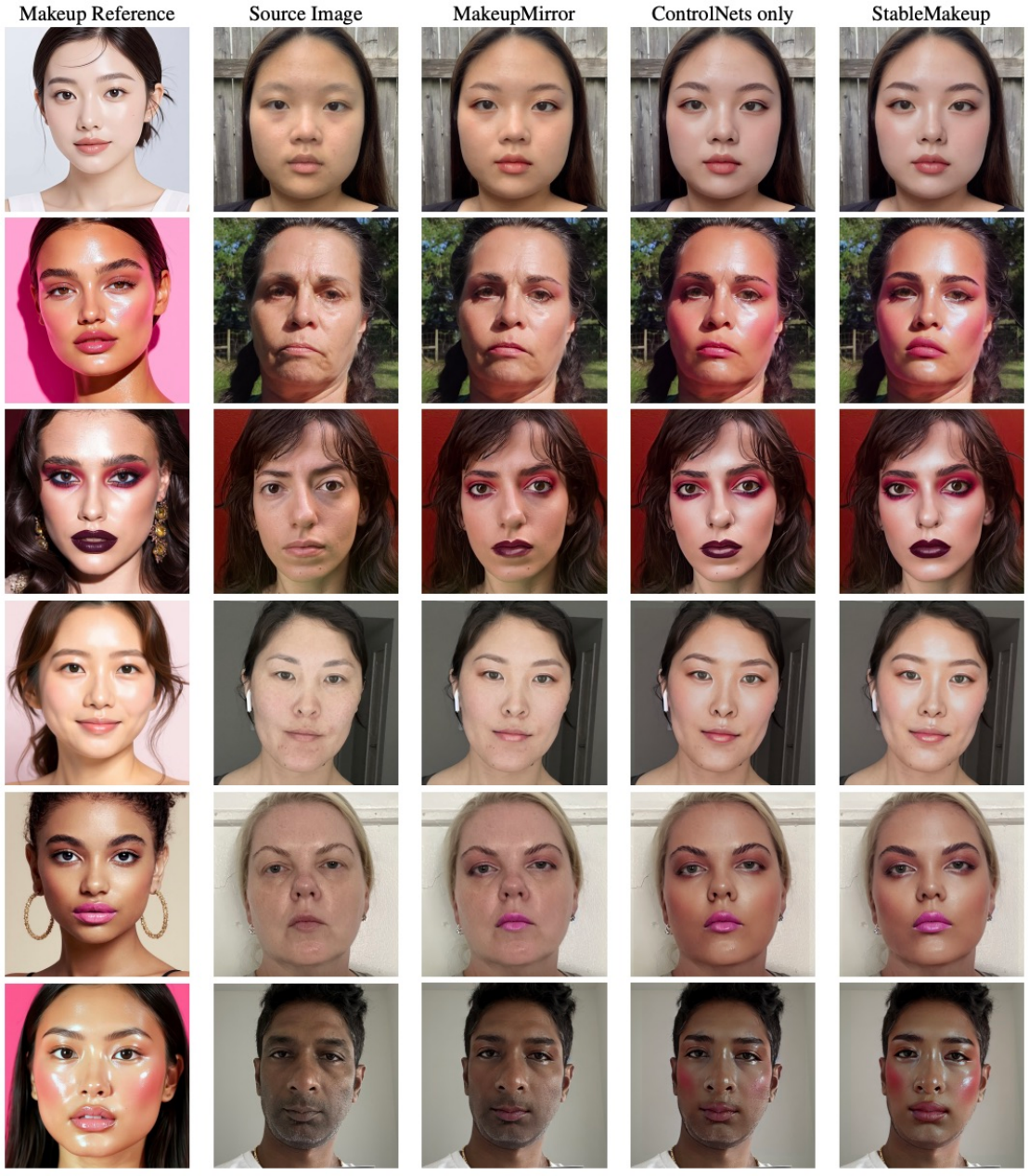}
  \caption{Qualitative results of each of our improvements compared to StableMakeup (column 5). StableMakeup suffers from changing facial geometry (\eg lip thickness, nose, eye shape, \etc) and skin tone, especially when copying makeup between different ethnicity. Our contributions alleviate these effects in large part.}
\label{fig:qualitative-results}
\end{figure*}

As we qualitatively see in \fig{qualitative-results}, our baseline, \ie Stable-Makeup, tends to change the thickness of lips, and geometry of eyes and nose.
While some of these changes might appear subtle, VTO applications are very personal experiences.
People easily notice subtle changes to their own facial geometry, making these methods not suitable for real-world applications.
By introducing depth- and edge-conditioned ControlNets, we significantly reduce these adverse qualitative effects.
More prominently, the baseline tends to copy the skin tone of the makeup reference image on the source face, resulting in unnatural looks which people would not want to wear in real-life.
Our second contribution addresses this problem, significantly reducing skin tone change and yielding more realistic makeup transfer. In a perceptual study, a low transfer threshold was generally found to produce the most favorable results. However, for makeup looks that include cheek-applied products such as blush, the adaptive skin tone transfer mechanism proposed in our work yields better perceptual outcomes. Notably, viewer tolerance for skin tone transfer intensity varies especially when input and output images are viewed side by side.

\begin{table}
\resizebox{\columnwidth}{!}{
\begin{tabular}{l|cccc}
\hline
Method &  Dv2 (M) & Dv2 (F) & $\Delta$ Monk & Face Sim. \\
\hline
Baseline & 
\textbf{0.583} & \textbf{0.627} & 1.132 & 0.487
\\
 + ControlNets & 
0.574 & 0.584 & 0.856 &  0.664 \\
 + Skin Tone & 
0.563 & 0.452 & \textbf{0.427} & \textbf{0.867} 
 \\
\hline
\end{tabular}
}
\caption{Our contributions consistently improve makeup transfer in terms of face similarity and monk scale difference, while preserving the makeup accuracy in makeup-specific regions.}
\label{tbl:ablation}
\end{table}

\subsection{Reducing Latency for Production}
\label{sec:latency}

Real-world VTO GenAI experiences for online shopping benefit from low latency and low operational cost. To this end, we reduce the inference time of our approach by leveraging recent advances in few-step inference approaches.
More specifically, we evaluate a Latent Consistency Models LoRA~(LCM)~\cite{luo2023latent} and a few step Levenberg-Marquardt-Langevin~(LML) sampler~\cite{wang2025unleashing}.

For a fair comparison, we used TensorRT (TRT) to speed up inference of the three variants of the model, and deploy them on NVIDIA L40S GPUs.
We summarize the latency, and number of model invocations (NFEs) of all approaches in Table~\ref{tbl:speedup}.
TRT already reduces the latency from 3.2s to 2.0s for the base model.
The LCM and LML optimizations bring significant speed-ups but also introduce approximations in inference and alter the outputs of the models.
To compare the quality of our methods, we performed a manual human audit on $500$ samples via pairwise preference scores, as well as measure their defects.
More specifically, we compare each method with our \methodname{} model and rate if its output is better (+1 points), equal ($\pm$0 points), or worse (-1 points) compared to \methodname{}.
Overall, the LML sampling method achieves slightly lower defects (3 / 500 vs 5 / 500) and higher preference compared to the LCM method ($\pm$ 0 / 500 vs -5 / 500), with a significant $2.8\times$ speed-up in inference time.

\begin{table}
\centering
\resizebox{\columnwidth}{!}{
\begin{tabular}{l|cccc}
    \hline
    \methodname{} & NFEs & Latency &  Preference & $\Delta$ Defects \\
    \hline
    with TensorRT     & 30    & 2.0s    & - & - \\  
    \hline
    with LCM+TRT & 4 & \textbf{0.45s}  & $- 5/500$  & +5/500 \\
    with LML+TRT & 10 & 0.70s  & $\mathbf{\pm 0 / 500}$ & \textbf{+3/500}  \\
    \hline
\end{tabular}
}
\caption{Evaluation of latency optimization approaches for our makeup transfer method. While not the fastest, using LML yields a $2.8\times$ speed-up without compromising on output quality.}
\label{tbl:speedup}
\end{table}

\section{Conclusion}
\label{sec:conclusion}

In this work, we present \methodname, a diffusion-based method for makeup transfer that marks a significant improvement in preserving skin tone and facial features from source images over Stable-Makeup.
These improvements are achieved by extending Stable-Makeup with three contributions: \begin{enumerate*} 
\item additional geometric conditioning in the ControlNet branch, namely with depth and low-level edge maps;
\item region-specific makeup strength control by leveraging face segmentation and adjusting transfer weight and clamped time-steps based on the region;
\item modulating the transfer strength to take into account the difference of skin tone between the reference and the source images.
\item leveraging Levenberg-Marquardt-Langevin sampling to achieve $2.8\times$ speed-up of the makeup transfer pipeline. 
\end{enumerate*}
Our experiments on public datasets and a newly collected one with more balanced skin tone distribution shows an improvement in relative facial similarity ($+60\%$) and reduction of skin tone alteration ($-60\%$) compared to Stable-Makeup, leading to a pass rate of $94\%$ according to an audit by beauty experts.
This high pass rate confirms the viability of the approach for a production-ready VTO experience for online beauty shopping.

{
    \small
    \bibliographystyle{ieeenat_fullname}
    \bibliography{main}

@String(CVPR= {IEEE Conf. Comput. Vis. Pattern Recog.})

@String(ECCV= {Eur. Conf. Comput. Vis.})

@String(NIPS= {Adv. Neural Inform. Process. Syst.})

@String(CVPR  = {CVPR})

@String(ECCV  = {ECCV})

@String(NIPS  = {NeurIPS})

@inproceedings{stable_makeup:2025,
author = {Zhang, Yuxuan and Yuan, Yirui and Song, Yiren and Liu, Jiaming},
title = {StableMakeup: When Real-World Makeup Transfer Meets Diffusion Model},
year = {2025},
isbn = {9798400715402},
publisher = {Association for Computing Machinery},
address = {New York, NY, USA},
url = {https://doi.org/10.1145/3721238.3730702},
doi = {10.1145/3721238.3730702},
booktitle = {Proceedings of the Special Interest Group on Computer Graphics and Interactive Techniques Conference Conference Papers},
articleno = {68},
numpages = {9},
series = {SIGGRAPH Conference Papers '25}
}

@inproceedings{zhang2023adding,
  title={Adding conditional control to text-to-image diffusion models},
  author={Zhang, Lvmin and Rao, Anyi and Agrawala, Maneesh},
  booktitle={Proceedings of the IEEE/CVF international conference on computer vision},
  pages={3836--3847},
  year={2023}
}

@inproceedings{yang2024depth,
  title={Depth anything: Unleashing the power of large-scale unlabeled data},
  author={Yang, Lihe and Kang, Bingyi and Huang, Zilong and Xu, Xiaogang and Feng, Jiashi and Zhao, Hengshuang},
  booktitle={Proceedings of the IEEE/CVF conference on computer vision and pattern recognition},
  pages={10371--10381},
  year={2024}
}

@article{cannyedge,
  author={Canny, John},
  journal={IEEE Transactions on Pattern Analysis and Machine Intelligence}, 
  title={A Computational Approach to Edge Detection}, 
  year={1986},
  volume={PAMI-8},
  number={6},
  pages={679-698},
  doi={10.1109/TPAMI.1986.4767851}}

@article{monk_2023, title={The Monk Skin Tone Scale}, url={osf.io/preprints/socarxiv/pdf4c_v1}, DOI={10.31235/osf.io/pdf4c}, publisher={SocArXiv}, author={Monk, Ellis}, year={2023}, month={May} }

@inproceedings{gu2019ladn,
  title={Ladn: Local adversarial disentangling network for facial makeup and de-makeup},
  author={Gu, Qiao and Wang, Guanzhi and Chiu, Mang Tik and Tai, Yu-Wing and Tang, Chi-Keung},
  booktitle={Proceedings of the IEEE/CVF International conference on computer vision},
  pages={10481--10490},
  year={2019}
}

@inproceedings{jiang2020psgan,
  title={PSGAN: Pose and expression robust spatial-aware gan for customizable makeup transfer},
  author={Jiang, Wentao and Liu, Si and Gao, Chen and Cao, Jie and He, Ran and Feng, Jiashi and Yan, Shuicheng},
  booktitle={Proceedings of the IEEE/CVF conference on computer vision and pattern recognition},
  pages={5194--5202},
  year={2020}
}

@inproceedings{Xiang:2022,
author = {Xiang, Jianfeng and Chen, Junliang and Liu, Wenshuang and Hou, Xianxu and Shen, Linlin},
title = {RamGAN: Region Attentive Morphing GAN for Region-Level Makeup Transfer},
year = {2022},
isbn = {978-3-031-20046-5},
publisher = {Springer-Verlag},
address = {Berlin, Heidelberg},
url = {https://doi.org/10.1007/978-3-031-20047-2_41},
doi = {10.1007/978-3-031-20047-2_41},
booktitle = {Computer Vision – ECCV 2022: 17th European Conference, Tel Aviv, Israel, October 23–27, 2022, Proceedings, Part XXII},
pages = {719–735},
numpages = {17},
location = {Tel Aviv, Israel}
}

@inproceedings{yang2022elegant,
  title={Elegant: Exquisite and locally editable gan for makeup transfer},
  author={Yang, Chenyu and He, Wanrong and Xu, Yingqing and Gao, Yang},
  booktitle={European conference on computer vision},
  pages={737--754},
  year={2022},
  organization={Springer}
}

@inproceedings{Li:2018,
author = {Li, Tingting and Qian, Ruihe and Dong, Chao and Liu, Si and Yan, Qiong and Zhu, Wenwu and Lin, Liang},
title = {BeautyGAN: Instance-level Facial Makeup Transfer with Deep Generative Adversarial Network},
year = {2018},
isbn = {9781450356657},
publisher = {Association for Computing Machinery},
address = {New York, NY, USA},
url = {https://doi.org/10.1145/3240508.3240618},
doi = {10.1145/3240508.3240618},
booktitle = {Proceedings of the 26th ACM International Conference on Multimedia},
pages = {645–653},
numpages = {9},
location = {Seoul, Republic of Korea},
series = {MM '18}
}

@inproceedings{ho2021classifier,
  title={Classifier-Free Diffusion Guidance},
  author={Ho, Jonathan and Salimans, Tim},
  booktitle={NeurIPS 2021 Workshop on Deep Generative Models and Downstream Applications},
    year={2021}
}

@inproceedings{mediapipe,
title	= {MediaPipe: A Framework for Perceiving and Processing Reality},
author	= {Lugaresi, Camillo and Tang, Jiuqiang and Nash, Hadon and McClanahan, Chris and Uboweja, Esha and Hays, Michael and Zhang, Fan and Chang, Chuo-Ling and Yong, Ming and Lee, Juhyun and Chang, Wan-Teh and Hua, Wei and Georg, Manfred and Grundmann, Matthias},
year	= {2019},
URL	= {https://mixedreality.cs.cornell.edu/s/NewTitle_May1_MediaPipe_CVPR_CV4ARVR_Workshop_2019.pdf},
booktitle	= {Third Workshop on Computer Vision for AR/VR at IEEE Computer Vision and Pattern Recognition (CVPR) 2019}}

@inproceedings{podellsdxl,
  title={SDXL: Improving Latent Diffusion Models for High-Resolution Image Synthesis},
  author={Podell, Dustin and English, Zion and Lacey, Kyle and Blattmann, Andreas and Dockhorn, Tim and M{\"u}ller, Jonas and Penna, Joe and Rombach, Robin},
  booktitle={The Twelfth International Conference on Learning Representations},year={2023}
}

@article{ramesh2022hierarchical,
  title={Hierarchical text-conditional image generation with clip latents},
  author={Ramesh, Aditya and Dhariwal, Prafulla and Nichol, Alex and Chu, Casey and Chen, Mark},
  journal={arXiv preprint arXiv:2204.06125},
  volume={1},
  number={2},
  pages={3},
  year={2022}
}

@inproceedings{rombach2022high,
  title={High-resolution image synthesis with latent diffusion models},
  author={Rombach, Robin and Blattmann, Andreas and Lorenz, Dominik and Esser, Patrick and Ommer, Bj{\"o}rn},
  booktitle={Proceedings of the IEEE/CVF conference on computer vision and pattern recognition},
  pages={10684--10695},
  year={2022}
}

@article{saharia2022photorealistic,
  title={Photorealistic text-to-image diffusion models with deep language understanding},
  author={Saharia, Chitwan and Chan, William and Saxena, Saurabh and Li, Lala and Whang, Jay and Denton, Emily L and Ghasemipour, Kamyar and Gontijo Lopes, Raphael and Karagol Ayan, Burcu and Salimans, Tim and others},
  journal={Advances in neural information processing systems},
  volume={35},
  pages={36479--36494},
  year={2022}
}

@inproceedings{Epstein:2023,
author = {Epstein, Dave and Jabri, Allan and Poole, Ben and Efros, Alexei A. and Holynski, Aleksander},
title = {Diffusion self-guidance for controllable image generation},
year = {2023},
publisher = {Curran Associates Inc.},
address = {Red Hook, NY, USA},
booktitle = {Proceedings of the 37th International Conference on Neural Information Processing Systems},
articleno = {714},
numpages = {18},
location = {New Orleans, LA, USA},
series = {NIPS '23}
}

@inproceedings{Li:2023,
author = {Li, Pengzhi and Huang, Qinxuan and Ding, Yikang and Li, Zhiheng},
title = {LayerDiffusion: Layered Controlled Image Editing with Diffusion Models},
year = {2023},
isbn = {9798400703140},
publisher = {Association for Computing Machinery},
address = {New York, NY, USA},
url = {https://doi.org/10.1145/3610543.3626172},
doi = {10.1145/3610543.3626172},
booktitle = {SIGGRAPH Asia 2023 Technical Communications},
articleno = {12},
numpages = {4},
location = {Sydney, NSW, Australia},
series = {SA '23}
}

@article{tsaban2023ledits,
  title={Ledits: Real image editing with ddpm inversion and semantic guidance},
  author={Tsaban, Linoy and Passos, Apolin{\'a}rio},
  journal={arXiv preprint arXiv:2307.00522},
  year={2023}
}

@article{xie2023edit,
  title={Edit everything: A text-guided generative system for images editing},
  author={Xie, Defeng and Wang, Ruichen and Ma, Jian and Chen, Chen and Lu, Haonan and Yang, Dong and Shi, Fobo and Lin, Xiaodong},
  journal={arXiv preprint arXiv:2304.14006},
  year={2023}
}

@inproceedings{zhang2023sine,
  title={Sine: Single image editing with text-to-image diffusion models},
  author={Zhang, Zhixing and Han, Ligong and Ghosh, Arnab and Metaxas, Dimitris N and Ren, Jian},
  booktitle={Proceedings of the IEEE/CVF conference on computer vision and pattern recognition},
  pages={6027--6037},
  year={2023}
}

@inproceedings{m_Nguyen-etal-CVPR21,
    author = {Thao Nguyen and Anh Tran and Minh Hoai},
    title = {Lipstick ain't enough: Beyond Color Matching for In-the-Wild Makeup Transfer},
    year = {2021},
    booktitle = {Proceedings of the {IEEE} Conference on Computer Vision and Pattern Recognition (CVPR)}
}

@article{sun2024shmt,
  title={SHMT: Self-supervised Hierarchical Makeup Transfer via Latent Diffusion Models},
  author={Sun, Zhaoyang and Xiong, Shengwu and Chen, Yaxiong and Du, Fei and Chen, Weihua and Wang, Fang and Rong, Yi},
  journal={Advances in neural information processing systems},
  year={2024}
}

@article{pan2026supervised,
  title={Supervised makeup transfer with a curated dataset: Decoupling identity and makeup features for enhanced transformation},
  author={Pan, Qihe and Wu, Yiming and Zhao, Xing and Xie, Liang and Sun, Guodao and Liang, Ronghua},
  journal={arXiv preprint arXiv:2602.00729},
  year={2026}
}

@article{serengil2026boosted,
  title     =  {{Boosted LightFace: A Hybrid DNN and GBM Model for Boosted Facial Recognition}},
  author    =  {Serengil, Sefik Ilkin and Ozpinar, Alper},
  journal   =  {Gazi University Journal of Science},
  year      =  {2026},
  doi       =  {10.35378/gujs.1794891},
  url       =  {https://dergipark.org.tr/en/pub/gujs/article/1794891},
  publisher =  {Gazi University}
}

@misc{oquab2023dinov2,
  title={DINOv2: Learning Robust Visual Features without Supervision},
  author={Oquab, Maxime and Darcet, Timothée and Moutakanni, Theo and Vo, Huy V. and Szafraniec, Marc and Khalidov, Vasil and Fernandez, Pierre and Haziza, Daniel and Massa, Francisco and El-Nouby, Alaaeldin and Howes, Russell and Huang, Po-Yao and Xu, Hu and Sharma, Vasu and Li, Shang-Wen and Galuba, Wojciech and Rabbat, Mike and Assran, Mido and Ballas, Nicolas and Synnaeve, Gabriel and Misra, Ishan and Jegou, Herve and Mairal, Julien and Labatut, Patrick and Joulin, Armand and Bojanowski, Piotr},
  journal={arXiv:2304.07193},
  year={2023}
}

@inproceedings{wang2025unleashing,
  title={Unleashing high-quality image generation in diffusion sampling using second-order levenberg-marquardt-langevin},
  author={Wang, Fangyikang and Yin, Hubery and Qian, Lei and Li, Yinan and Zhuang, Shaobin and Zhu, Huminhao and Zhang, Yilin and Tang, Yanlong and Zhang, Chao and Zhao, Hanbin and others},
  booktitle={Proceedings of the IEEE/CVF International Conference on Computer Vision},
  DELpages={10453--10464},
  year={2025}
}

@misc{luo2023latent,
  title={{Latent Consistency Models: Synthesizing High-Resolution Images with Few-Step Inference}},
  author={Simian Luo and Yiqin Tan and Longbo Huang and Jian Li and Hang Zhao},
  year={2023},
  eprint={2310.04378},
  archivePrefix={arXiv},
  primaryClass={cs.CV}
}

@article{lu2024makeupdiffuse,
  title={MakeupDiffuse: a double image-controlled diffusion model for exquisite makeup transfer},
  author={Lu, Xiongbo and Liu, Feng and Rong, Yi and Chen, Yaxiong and Xiong, Shengwu},
  journal={The Visual Computer},
  pages={1--17},
  year={2024},
  publisher={Springer}
}

@article{zhu2025flux,
  title={FLUX-Makeup: High-Fidelity, Identity-Consistent, and Robust Makeup Transfer via Diffusion Transformer},
  author={Zhu, Jian and Liu, Shanyuan and Li, Liuzhuozheng and Gong, Yue and Wang, He and Cheng, Bo and Ma, Yuhang and Wu, Liebucha and Wu, Xiaoyu and Leng, Dawei and others},
  journal={arXiv preprint arXiv:2508.05069},
  year={2025}
}

@article{park2025dreammakeup,
  title={DreamMakeup: Face Makeup Customization using Latent Diffusion Models},
  author={Park, Geon Yeong and Han, Inhwa and Yang, Serin and Hong, Yeobin and Jeong, Seongmin and Jeon, Heechan and Goh, Myeongjin and Yi, Sung Won and Nam, Jin and Ye, Jong Chul},
  journal={arXiv preprint arXiv:2510.10918},
  year={2025}
}

@article{Tong2007ExampleBasedCT,
  title={Example-Based Cosmetic Transfer},
  author={Tong, Wai-Shun and Tang, Chi-Keung and Brown, M. S. and Xu, Ying-Qing},
  journal={15th Pacific Conference on Computer Graphics and Applications (PG'07)},
  year={2007},
  pages={211-218},
  url={https://api.semanticscholar.org/CorpusID:11242938}
}

@article{Guo2009DigitalFM,
  title={Digital face makeup by example},
  author={Dong Guo and Terence Sim},
  journal={2009 IEEE Conference on Computer Vision and Pattern Recognition},
  year={2009},
  pages={73-79},
  url={https://api.semanticscholar.org/CorpusID:11904491}
}

@inproceedings{Lin:2013,
author = {Xu, Lin and Du, Yangzhou and Zhang, Yimin},
year = {2013},
month = {09},
pages = {3206-3210},
title = {An automatic framework for example-based virtual makeup},
booktitle = {Proceedings of the 20th IEEE International Conference on Image Processing},
doi = {10.1109/ICIP.2013.6738660}
}

@article{Liu:2024,
author = {Liu, Luoqi and Xing, Junliang and Liu, Si and Xu, Hui and Zhou, Xi and Yan, Shuicheng},
title = {“Wow! You Are So Beautiful Today!”},
year = {2014},
issue_date = {September 2014},
publisher = {Association for Computing Machinery},
address = {New York, NY, USA},
volume = {11},
number = {1s},
issn = {1551-6857},
url = {https://doi.org/10.1145/2659234},
doi = {10.1145/2659234},
journal = {ACM Trans. Multimedia Comput. Commun. Appl.},
month = oct,
articleno = {20},
numpages = {22},
}

@article{Li:2019,
author = {Li, Chen and Zhou, Kun and Wu, Hsiang-Tao and Lin, Stephen},
title = {Physically-Based Simulation of Cosmetics via Intrinsic Image Decomposition with Facial Priors},
year = {2019},
issue_date = {June 2019},
publisher = {IEEE Computer Society},
address = {USA},
volume = {41},
number = {6},
issn = {0162-8828},
url = {https://doi.org/10.1109/TPAMI.2018.2832059},
doi = {10.1109/TPAMI.2018.2832059},
journal = {IEEE Trans. Pattern Anal. Mach. Intell.},
month = jun,
pages = {1455–1469},
numpages = {15}
}

@article{Blattmann2023StableVD,
  title={Stable Video Diffusion: Scaling Latent Video Diffusion Models to Large Datasets},
  author={A. Blattmann and Tim Dockhorn and Sumith Kulal and Daniel Mendelevitch and Maciej Kilian and Dominik Lorenz},
  journal={ArXiv},
  year={2023},
  volume={abs/2311.15127},
  url={https://api.semanticscholar.org/CorpusID:265312551}
}

@article{mou2023t2i,
  title={T2I-Adapter: Learning Adapters to Dig out More Controllable Ability for Text-to-Image Diffusion Models},
  author={Mou, Chong and Wang, Xintao and Xie, Liangbin and Zhang, Jian and Qi, Zhongang and Shan, Ying and Feng, Xiaohu},
  journal={arXiv preprint arXiv:2302.08453},
  year={2023}
}

@article{yu2023lora,
  title={LoRA: Low-Rank Adaptation for Fast Text-to-Image Diffusion Fine-tuning},
  author={Yu, Haonan and Chen, Xiangyu and Chen, Kunhao and Shi, Weiwei and Xie, Xiaodong and Zhang, Yong and Qin, Tao and Liu, Tie-Yan},
  journal={arXiv preprint arXiv:2307.02904},
  year={2023}
}

@article{wu2022unified,
  title={A Unified Framework for Incorporating Human Feedback into Text-to-Image Generation},
  author={Wu, Yichi and Zhang, Hongming and Li, Xiaohui and Wen, Nian and Gao, Yingxue and Shen, Yelong and Duan, Nan},
  journal={arXiv preprint arXiv:2305.06500},
  year={2023}
}

@inproceedings{wallace2024diffusiondpo,
  title={Diffusion Model Alignment Using Direct Preference Optimization},
  author={Wallace, Bram and others},
  booktitle={Proceedings of the IEEE/CVF Conference on Computer Vision and Pattern Recognition (CVPR)},
  year={2024},
  url={https://arxiv.org/abs/2311.12908}
}

@article{ho2020denoising,
  title={Denoising Diffusion Probabilistic Models},
  author={Ho, Jonathan and Jain, Ajay and Abbeel, Pieter},
  journal={Advances in Neural Information Processing Systems},
  volume={33},
  pages={6840--6851},
  year={2020}
}

@article{dhariwal2021diffusion,
    title={Diffusion Models Beat GANs on Image Synthesis},
    author={Dhariwal, Prafulla and Nichol, Alexander},
    journal={Advances in Neural Information Processing Systems},
    volume={34},
    year={2021}
}

@article{kling2025klingomni,
  title={Kling-Omni: A Generalist Generative Framework for Multimodal Video Synthesis},
  author={Kling Team},
  journal={arXiv preprint arXiv:2512.16776},
  year={2025},
  url={https://arxiv.org/abs/2512.16776}
}

@article{wan2024videogen,
  title={Wan: Open and Advanced Large-Scale Video Generative Models},
  author={Wan Team},
  journal={arXiv preprint arXiv:2406.09203},
  year={2024},
  url={https://arxiv.org/abs/2406.09203}
}

@article{lugmayr2022repaint,
  title={RePaint: Inpainting using Denoising Diffusion Probabilistic Models},
  author={Lugmayr, Andreas and Danelljan, Martin and Romero, Andres and Yu, Fisher and Timofte, Radu and Van Gool, Luc},
  journal={arXiv preprint arXiv:2201.09865},
  year={2022},
  url={https://arxiv.org/abs/2201.09865}
}

@inproceedings{Radford2021LearningTV,
  title={Learning Transferable Visual Models From Natural Language Supervision},
  author={Alec Radford and Jong Wook Kim and Chris Hallacy and A. Ramesh and Gabriel Goh and Sandhini Agarwal and Girish Sastry and Amanda Askell and Pamela Mishkin and Jack Clark and Gretchen Krueger and Ilya Sutskever},
  booktitle={ICML},
  year={2021}
}
}


\end{document}